\documentclass[conference]{IEEEtran}
\IEEEoverridecommandlockouts
\usepackage{cite}
\usepackage{amsmath,amssymb,amsfonts}
\usepackage{algorithmic}
\usepackage{graphicx}
\usepackage{textcomp}
\usepackage{xcolor}

\usepackage[ruled,norelsize]{algorithm2e}
\newtheorem{theorem}{Theorem}

\makeatletter
\newcommand{\removelatexerror}{\let\@latex@error\@gobble}
\makeatother

\def\BibTeX{{\rm B\kern-.05em{\sc i\kern-.025em b}\kern-.08em
    T\kern-.1667em\lower.7ex\hbox{E}\kern-.125emX}}
\begin{document}

\title{Regularized Sequential Latent Variable Models with Adversarial Neural Networks}

\author{

\IEEEauthorblockN{1\textsuperscript{st} Jin Huang}
\IEEEauthorblockA{\textit{the Department of Information Science and Engineering} \\
\textit{KTH Royal Institute of Technology}\\
Stockholm, Sweden \\
huangj@kth.se}
\and

\IEEEauthorblockN{2\textsuperscript{ed} Ming Xiao}
\IEEEauthorblockA{\textit{the Department of Information Science and Engineering} \\
	\textit{KTH Royal Institute of Technology}\\
	Stockholm, Sweden \\
	mingx@kth.se}

}
\maketitle

\begin{abstract}
	The recurrent neural networks (RNN) with richly distributed internal states and flexible non-linear transition functions, have overtaken the dynamic Bayesian networks such as the hidden Markov models (HMMs) in the task of modeling highly structured sequential data. These data, such as from speech and handwriting, often contain complex relationships between the underlaying variational factors and the observed data.  The standard RNN model has very limited randomness or variability in its structure, coming from the output conditional probability model. This paper will present different ways of using high level latent random variables in RNN to model the variability in the sequential data, and the training method of such RNN model under the VAE (Variational Autoencoder) principle. We will explore possible ways of using adversarial method to train a variational RNN model. Contrary to competing approaches, our approach has theoretical optimum  in the model training and provides better model training stability. Our approach also improves the posterior approximation in the variational inference network by a separated adversarial training step. Numerical results simulated from TIMIT speech data show that reconstruction loss and evidence lower bound converge to the same level and adversarial training loss converges to 0. 
\end{abstract}


\section{Introduction}\label{sec:introduction}

The sequential data modeling, especially on the high dimensional data such as speech and music data, has been a challenge in machine learning. Real world applications often require the sequential model to learn the true data distributions and generate data points with variations, i.e, a generative sequential model. Historically, Dynamic Bayesian Networks (DBNs) such Hidden Markov Model (HMM) have been widely studied and used for sequential data modeling. Training these models do not require optimization of model parameters in the high dimensional space. As the available computing power increases, there is a resurgence of interest in using recurrent neural networks for modeling the sequential data. The original RNN has overturned DBNs in modeling the sequential data. However, the original RNN has the deterministic state transition structure, comparing to the random variable hidden state in DBNs. As the sequential data in the real world often come with random transition between the adjacent underling states of the observations, introducing more randomness to RNN has been widely studied. There are different ways of introducing extra randomness to the original RNN structure.  Latent variables are the hidden variables in the sequential model with certain probability distribution. Introducing latent variable to original RNN is proven to be an efficient way to introduce extra randomness and improve the performance for RNN to learn the distribution of the sequential data. 

Variational Recurrent Neural Network (VRNN) is a sequential latent variable model that includes latent states in the transition structure of the deterministic RNN \cite{chung2015recurrent}. Due to the introduced random variables in the RNN, training the network with gradient descent on the log likelihood objective function becomes difficult as the objective function becomes intractable. Estimators based on different variational bounds for the sequential latent variable model have been proposed and studied. Evidence Lower Bound (ELBO) has produced state of the art results in maximum likelihood estimation (MLE) for the latent variable model with a sequential structure. Important Weighted Autoencoders (IWAE), and filtering variational objectives (FIVOs) are extensions of ELBO, defined by the particle filter’s estimator of the marginal log likelihood on sequential observations.

ELBO comes from the study on variational autoencoder (VAE) \cite{kingma2013auto} \cite{kingma2019introduction}, and they then are extended to sequential latent variable models.  VAE has emerged as a popular approach for learning on complicated data distributions. Sepcifically, let x denote the observation of random variable in a high dimensional space $\mathcal{X}$, and $z$ denote a latent random variable involved in the process of generating the observations. VAE maximizes a lower bound, i.e. ELBO, of marginal log likelihood $log p(x)$ as below:
\begin{equation}
	log p(x) \geq -KL(q(z|x)\parallel p(z))+\mathbb{E}_{q(z|x)}[log p(x|z)]. 
\end{equation} 

Here $q(z | x)$  is a variational approximation of the posterior $p(z|x)$. We can interpret this lower bound as reconstruction and regulation. The second item of this bound is reconstruction of $p(x)$ from the approximated posterior distribution on the latent variable $z$, and the other item with Kullback-Leibler (KL) divergence regularizes the reconstructed distribution by imposing the prior distribution $p(z)$ on the inference posterior distribution $q(z|x) $.

ELBO is a variational objective taking a variational posterior $q$ as an argument. In \cite{maddison2017filtering}, the important question that whether variational bounds like ELBO achieve the marginal log-likelihood at its optimal $q$ is studied. It shows that gradient decent optimization on the variational bounds can not reconstruct the marginal log-likelihood on data observations at the optimal $q$. The sharpness of ELBO is not guaranteed by existing methods. With these consideration, the performance of state of the art latent variable models can be improved by a better regularization method.

ELBO has been extensively used in  sequential latent variable models such as VRNN, by introducing factorization on the sequence distributions. Our work focuses on improving the regularization of the reconstructed sequential data distribution by introducing adversarial neural networks in the model. We propose a model called Adversarial Regularized Variational RNN (A-VRNN) that achieves the optimal state at its optimal posterior approximation $q$. This has not been achieved by existing methods. For experiments, we train the A-VRNN model using speech sequential data, and show superior performance.

\section{Background}
\subsection{Sequence modeling with Recurrent Neural Networks}

RNN is a family of neural network structures that are widely used in modeling sequential data. It has a basic cell unit that takes the sequence input $x=(x_1, x_2, ..., x_T)$ recursively. Each data point is processed to with the same set to cell trainable parameters while maintaining its internal hidden state $h$. At each time-step $t$, RNN reads $x_t$ and updates its hidden state $h_t$ and produces and output based on the new hidden state. $h_t$ is updated by

\begin{equation}
	h_t=f_\theta(x_t,h_{t-1}),
\end{equation} 

where f is a deterministic non-linear transition function represented by the RNN cell unit, and $\theta$ is the parameter set of the cell unit. The state of the art RNN cells such as long short-term memory (LSTM) and gated recurrent unit (GRU) are implemented with gated activation functions \cite{hochreiter1997long}, \cite{cho2014learning}. An output function maps the RNN hidden state $h_{t}$ to a probability distribution of the output data:

\begin{equation}
	p(x_t|x_{<t})=g(h_{t-1})),
\end{equation} 

where $g$ is the output function represented by output gate in the RNN cell and returns the parameters of the output data distribution, conditioned on $h$. The joint sequence probability distribution is factorized by a product of these conditional probability as:

\begin{equation}
	p(x_1,x_2,...,x_T)=\prod_{t=1}^{T}p(x_t|x_{<t}).
\end{equation} 
The choice of output distribution models depends on the observations. Gaussian mixture model (GMM) is a common choice for modeling high-dimensional sequences. The internal transition of original RNN models is entirely deterministic and the variability and randomness of the observed sequences is modeled by the conditional output probability density.

The need for introducing more variability to original RNN models has been previously noted. Sequential latent variable models (SLVM) based on RNN such as STORN and VRNN have proved to produce better performance for modeling highly variable sequential data such as speech and music \cite{chung2015recurrent}, \cite{bayer2014learning}.  These RNN based SLVM introduces latent variables $z=(z_1,z_2,...,z_T)$ to the transition structure. At each time-step $t$, the transition function $f$ computes next the hidden state $h_t$ based on both the previous hidden state $h_{t-1}$ and the latent variable $z_t$. STORN and VRNN have different ways of generating latent variable $z$. STORN models $z$ as a sequence of independent random variables, while VRNN models the prior distribution $z_t$ dependent on all the preceding inputs $(x_1,x_2,...,x_{t-1})$ via the previous RNN hidden state $h_{t-1}$. VRNN can be interpreted as containing a VAE at every time-step. The generative model of VRNN is factorized as:

\begin{equation} \label{generation}
	p(x_{\leq T},z_{\leq T})=\prod_{t=1}^{T}p(x_t|z_{\leq t},x_{<t})p(z_t|x_{<t},z_{<t}).
\end{equation} 

The inference of the approximate posterior in VRNN is factorized as:

\begin{equation}\label{inference}
	q(z_{\leq T}|x_{\leq T})=\prod_{t=1}^{T}q(z_t|x_{\leq t},z_{<t}).
\end{equation}

\subsection{Evidence Lower Bound in sequential model}
The approach of training RNN models with latent variables is inspired and similar as in the standard VAE. If we look at each time-step of VRNN models, it includes several operations:

\begin{itemize}
	\item Compute the prior distributions of $z_t$ conditioned on previous hidden state $h_{t-1}$. This operation is also called 'transition'.
	\item Generate $x_t$ based on generating distribution conditioned on both $z_t$ and $h_{t-1}$. This operation is also called 'emission'.
	\item Update the hidden state by running the deterministic RNN for on step, taking previous hidden state $h_{t-1}$, latent state $z_{t-1}$ and observation $x_t$. 
	
	\item Run inference of the approximate posterior distribution of $z_t$ using the mean from the prior and taking $x_t$ and $h_{t-1}$ as input. This operation is also called 'proposal'.
\end{itemize}

Just like in VAE, the generating (emission) and inference (proposal) networks are trained jointly by maximizing the variational lower bound with  respect to their parameters \cite{chung2015recurrent}. By using concavity and Jansen's inequality on the joint log probability of the whole data sequence, we have:

\begin{equation}
	log p(x_{1:T})= log \int \frac{p(x_{\leq T},z_{\leq T})}{q(z_{\leq T}|x_{\leq T})} q(z_{\leq T}|x_{\leq T}) dz,
\end{equation}

\begin{equation}\label{elbo1}
	log p(x_{1:T})\geq \mathbb{E}_{q(z_{\leq T}|x_{\leq T})} [log \frac{p(x_{\leq T},z_{\leq T})}{q(z_{\leq T}|x_{\leq T})}].
\end{equation}

Using the factorization in (\ref{generation}) and (\ref{inference}), we have:

\begin{equation}
	log p(x_{1:T})\geq \mathbb{E}_{q(z_{\leq T}|x_{\leq T})} [log \prod_{t=1}^{T} \frac{p(x_t|z_{\leq t},x_{<t})p(z_t|x_{<t},z_{<t}) }{q(z_t|x_{\leq t},z_{<t})}].
\end{equation} 

The ELBO in variational RNN becomes factorized variational lower bound:

\begin{multline}  \label{elbo}
\mathbb{E}_{q(z_{\leq T}|x_{\leq T})}[\sum_{t=1}^{T}(logp(x_t|z_{\leq t},x_{<t})\\
-KL(q(z_t|x_{\leq t},z_{<t})||p(z_t|x_{<t},z_{<t})))].
\end{multline}

Similarly, this bound has a reconstruction item $\sum_{t=1}^{T} logp(x_t|z_{\leq t},x_{<t})$ and a regularization item in the form of Kullback-Leibler divergence \cite{kullback1997information}.

\section{Related Work} \label{related}

As aforementioned, the early work of using latent variable on generative models is VAE \cite{kingma2013auto}. This approach has been successful on solving unsupervised learning and semi-supervised learning problems \cite{walker2016uncertain}, \cite{abbasnejad2017infinite}. Different encoder and decoder neural networks in VAE have been studied to improve the performance in these problems \cite{pu2016variational}. Related works in VAE show its limitation in approximation of the posterior distributions. Reference \cite{burda2015importance} uses multiple samples to approximate the posterior, giving it increased flexibility to model complex posteriors, which improve the capability of VAE in modeling complex distributions. In  \cite{sonderby2016ladder}, a data dependent approximate
likelihood is studied to correct the generative distribution.

RNN has been widely used in the sequential data modeling, and the latent variable approach is extended to sequential generative model with RNN as base models \cite{chung2015recurrent,maddison2017filtering,fraccaro2016sequential}. The latent variables bring additional variance and randomness to RNN, and make the model more expressive. Different ways of using RNN for structured variational approximation in latent variable sequential models are studied in \cite{krishnan2017structured}. The difference is mainly about the dependency of conditional posterior distributions.

After GAN was introduced to train generative models, the GAN model has been adopted and improved in related research works \cite{goodfellow2014generative}. To use GAN in sequential model training, the generator model generally need to be a conditional distribution model. In \cite{mirza2014conditional}, the conditional version of GAN is proposed with conditions in both generator and discriminator model. In our approach, the generator models in the adversarial training will be conditional distributions depending on the observations and latent variables. Some other works in the area of GAN combine adversarial training with Convectional Neural Networks (CNN), demonstrating its applicability in image data representations \cite{denton2015deep}, \cite{radford2015unsupervised}. 

Adversarial training on VAE is studied in related works like \cite{makhzani2015adversarial},\cite{mescheder2017adversarial}, with combination of VAE and GAN. The model proposed in \cite{makhzani2015adversarial} is called Adversarial Auto-Encoder (AAE), which performs variational inference by matching the aggregated posterior of the hidden code vector of the autoencoder with an arbitrary prior distribution. However, these approaches are not derived for sequential latent variable models.

Our contributions to this research area are:

\begin{enumerate}
	\item We propose a novel approach of regularizing sequential latent variable model, with adversarial training.
	\item We prove our regularization approach achieves theoretical optimum when training a sequential latent variable model. This improves the model sharpness comparing to using ELBO as the training objective.
	\item We prove the equivalence of the reconstruction loss and the Evidence Lower Bound when the adversarial training achieves optimum.
	\item Our theoretical analysis proves that when the discriminator is at its optimum, adversarial training objective is Earth-Mover distance. This shows our regularization approach makes the training smooth and stable.
	\item We propose a novel approach with separated optimization steps for the auto-encoder and latent distribution models. This gives clear tracks on the factorization of the posterior distribution.

\end{enumerate}

\section{Adversarial regularized variational RNN (A-VRNN)} \label{adv}
\begin{figure*}[!h]
	\centering
	\fbox{\includegraphics[height=7.5cm]{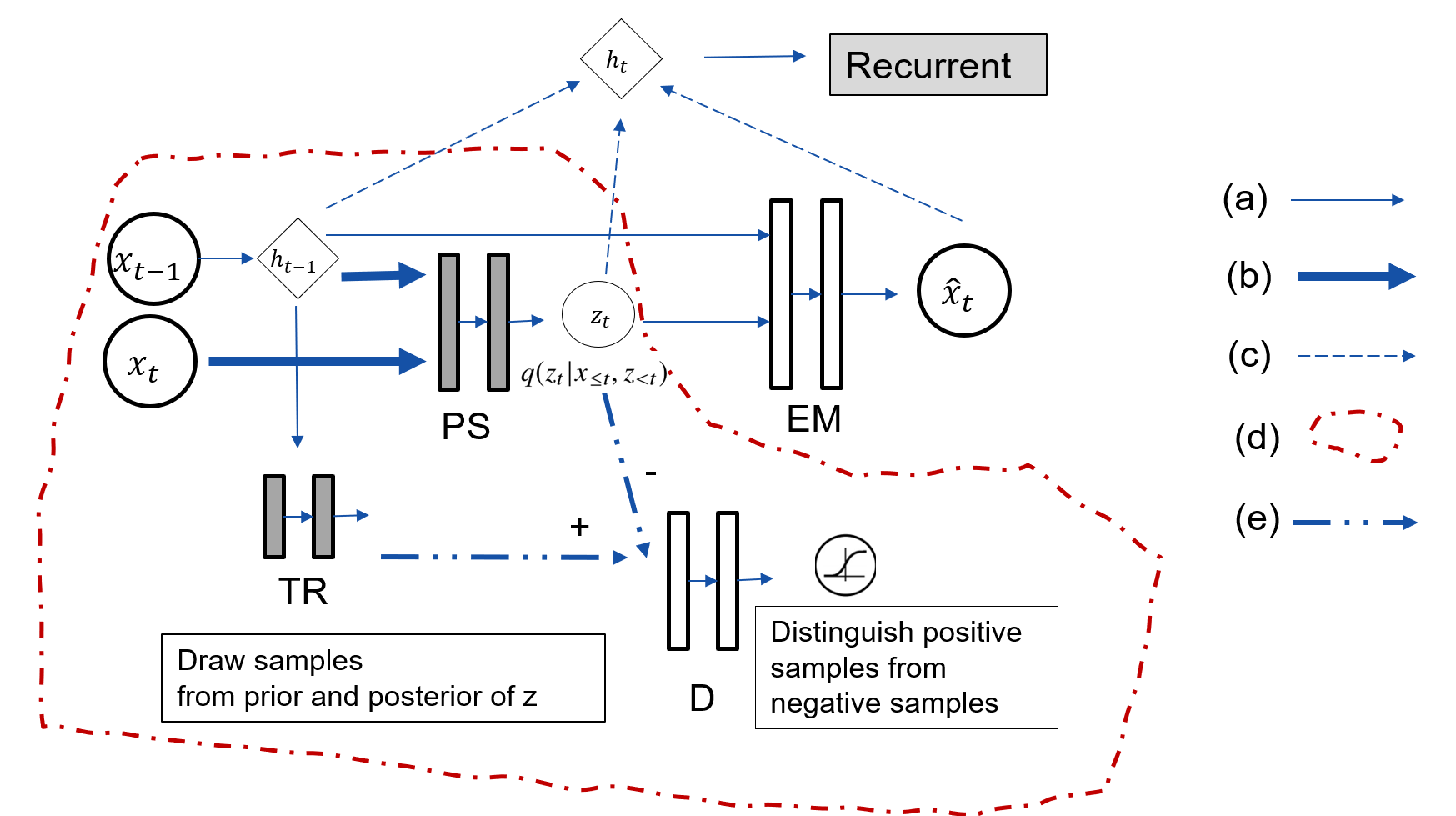}}
	
	\caption{Graphical illustrations of A-VRNN: (a) Forward steps and general connection; (b) Inference network that computes approximate posterior; (c) Update the hidden state of RNN; (d) The adversarial regularization; (e) Prior and Posterior sampling of $z_t$. Neural networks models in A-VRNN: TR (Transition Model), PS (Proposal Model), EM (Emission Model), D (Discriminator Model).}
	\label{fig:model}
\end{figure*}

In this section, we introduce a new version of sequential latent variable model A-VRNN, regularized by adversarial neural networks. A-VRNN keeps the flexibility of the original VRNN for modeling highly non-linear sequential dynamics, while providing smoother and sharper distance measurement for regularization on latent prior and posterior distribution \cite{weng2019gan}. In our proposed model, the optimization on the variational bound become separate training steps, reconstruction and regularization. The regularization training also has separate steps, discriminator training and adversarial training. This brings benefit for model training stability and improves the posterior approximation in the inference network \cite{fraccaro2016sequential}.

Recall that in the latent variable sequential model, we have observation $x=(x_1,x_2,...,x_T)$. A-VRNN model takes the observation data as input once at time, and updates its hidden state at each time step. The output of each time-step is a probability distribution conditioned on a latent variable $z_t$ and current hidden state. In A-VRNN, an discriminator neural network is added in the structure to play mini-max game with the inference network. The discriminator takes samples from both proposal model $q(z_t|z_{<t}, x_{\leq t})$ and transition model $p(z_t|x_{<t},z_{<t})$. Note that in these two models, past $x_{<t}$ and $z_{<t}$ are connected indirectly with current $z_t$ by conditional probability model based on $h$. Our proposed model is illustrated in Fig. \ref{fig:model}. The discriminator $D$ outputs $1$ when it believes the sample is from transition model output prior distribution of $z_t$, and outputs $0$ when it believes the sample is from proposal model approximate posterior distribution. 

The transition model (TR) is a conditional Gaussian distribution parameterized by neural networks. TR computes the prior distribution $p(z_t|h_{t-1})$.  The emission model (EM) is a conditional Bernoulli distribution, also parametrized by neural networks. EM computes the output target distribution $p(x_t|z_t, h_{t-1})$. $\hat{x_t}$ is the sample drawn from the output distribution. The proposal model (PS) is a conditional Gaussian distribution. PS computes the approximate posterior based on observation $x_t$ and previous hidden state. 

The general objective of training this sequential latent variable model is to learn the data distribution based on the observed training data. The reconstruction phase when training A-VRNN is to maximize the reconstructed log likelihood of $x$ given the approximate posterior of $z$. The factorized reconstruction training can be expressed as:

\begin{equation}
	\displaystyle \max \mathbb{E}_{q(z_{\leq T}|x_{\leq T})}\sum_{t=1}^{T}\left[\log P (x_t|x_{< t},z_{\leq t})\right].
\end{equation}

The regularization phase is a mini-max game to impose a regularization on the approximated posterior of $z$ with a prior distribution conditioned on previous data observations and latent variables through $h_{t-1}$. The regularization phase of the training can be expressed as:

\begin{multline}  
	\displaystyle \min_{PS}\max_D\sum_{t=1}^{T}[\mathbb{E}_{p(z_t|x_{< t},z_{< t})} \log D(z_t) \\ 
	+\mathbb{E}_{x \sim p_{data}}\log[1-D(PS(x_{\leq t})]].
\end{multline}

Algorithm \ref{alg1} and Algorithm \ref{alg2} describe our proposed learning algorithm. In the adversarial training part, we use Wasserstein GAN (WGAN) instead of the original GAN method \cite{arjovsky2017wasserstein}. We use RMSProp optimizer as recommened in WGAN \cite{tieleman2012rmsprop}. $f_\theta()$ is the deterministic function of RNN cell, and $\theta$ is the trainable parameters of the neural networks. $TR_\omega(),EM_\phi(),PS_\tau()$ are highly flexible functions parameterized by neural networks. These three functions compute the parameters of the conditional probability distribution in 'Transition', 'Emission', and 'Proposal'. $\omega,\phi,\tau$ are the trainable parameters in these three neural networks. $D_\eta()$ is the discriminator model represented by a neural networks that computes the probability that a sample of $z$ come from prior distribution by $TR_\omega()$ (positive sample), rather than from posterior distribution by $PS_\tau()$. $\eta$ is the trainable parameters in discriminator model.

\begin{figure}[!t] 
	\removelatexerror
	\begin{algorithm}[H] \label{alg1}
		\caption{Our proposed algorithm for training a sequential latent variable model. $\bar{x}^{(i)}=[x_1^{(i)},...,x_T^{(i)}]$ represent the observed data samples in batch. $B$ is number of training iteration; $m$ is batch size of train data; $g$ represent the gradient decent for the trainable parameters.}
		\For {$B$ training iterations}
		{
			\textbf{(1) Reconstruction phase}
			
			Sample $\left\lbrace \bar{x}^{(i)}\right\rbrace  _{i=1}^m \sim P_{data}$, a batch from training data.
			
			\For {$t=0,...,T$}
			{
				Compute the prior probablity distribution $P(z_t) \Leftarrow TR_\omega (h_{t-1})$.
				
				Compute the posterior probablity distribution $Q(z_t) \Leftarrow PS_\tau (\bar{x_t},h_{t-1})$.	
				
				Sample $\tilde{z_t} \sim Q(z_t)$.
				
				Compute the output probablity distribution $P(x_t|z_t) \Leftarrow EM_\phi (\tilde{z_t},h_{t-1})$.
				
				$log p(\left\lbrace \bar{x}_t^{(i)}\right\rbrace) = logP(x_t|z_t)|_{x_t=\left\lbrace \bar{x}_t^{(i)}\right\rbrace}$
				
				Compute recontstuction loss for current step: $\mathcal{L}_{rec}^t =-\sum_{i=1}^{m} log p(\left\lbrace \bar{x}_t^{(i)}\right\rbrace)$
				
				Update hidden state $h_t \leftarrow f_\theta (h_{t-1},\left\lbrace \bar{x}_{t-1}^{(i)}\right\rbrace)$
				
			}
			$\mathcal{L}_{rec}=\sum_{t=1}^{T}\mathcal{L}_{rec}^t$
			
			$g_\theta,g_\phi,g_\tau \leftarrow \nabla_{\theta,\phi,\tau} \mathcal{L}_{rec}$
			
			$\theta,\phi,\tau \leftarrow RMSPropg(g_\theta,g_\phi,g_\tau)$

			\textbf{(2) Regularization phase}
			
			Adversarial regularization using Algorithm \ref{alg2}.
			
		}
		
	\end{algorithm}
\end{figure}

\begin{figure}[!t]
	\removelatexerror
	\begin{algorithm}[H] \label{alg2}
		\caption{One iteration of our proposed Adversarial Regularization Algorithm.}

		\textbf{(1) Discriminator training}
		
		Sample $\left\lbrace \bar{x}^{(i)}\right\rbrace  _{i=1}^m \sim P_{data}$, a batch from training data.
		
		\For {$t=0,...,T$}
		{
			Compute the prior probablity distribution $P(z_t) \Leftarrow TR_\omega (h_{t-1})$.
			
			Compute the posterior probablity distribution $Q(z_t) \Leftarrow PS_\tau (\bar{x_t},h_{t-1})$.	
			
			Sample $\left\lbrace z_t^{(i)}\right\rbrace \sim P(z_t)$.
			
			Sample $\left\lbrace \tilde{z}_t^{(i)}\right\rbrace\sim Q(z_t)$.
			
			Update hidden state $h_t \leftarrow f_\theta (h_{t-1},\left\lbrace \bar{x}_{t-1}^{(i)}\right\rbrace)$
			
		}
		
		$\left\lbrace z^{(i)}\right\rbrace = [\left\lbrace z_1^{(i)}\right\rbrace,...,\left\lbrace z_T^{(i)}\right\rbrace  ]$;
		$\left\lbrace \tilde{z}^{(i)}\right\rbrace=[\left\lbrace \tilde{z}_1^{(i)}\right\rbrace,...,\left\lbrace \tilde{z}_T^{(i)}\right\rbrace]$.
		
		Compute the discrimination loss: $\mathcal{L}_{dis}=\sum_{i=1}^{m} [D_\eta(z^{(i)})-D_\eta(\tilde{z}^{(i)})]$
		
		$g_\eta \leftarrow \nabla_\eta \mathcal{L}_{dis}$
		
		$\eta \leftarrow RMSProp (g_\eta) $
		
		$\eta \leftarrow clip (\eta,-c,c)$

		\textbf{(2) Adversarial training}
		
		Sample $\left\lbrace \bar{x}^{(i)}\right\rbrace  _{i=1}^m \sim P_{data}$, a batch from training data.
		
		\For {$t=0,...,T$}
		{
			Compute the prior probablity distribution $P(z_t) \Leftarrow TR_\omega (h_{t-1})$.
			
			Compute the posterior probablity distribution $Q(z_t) \Leftarrow PS_\tau (\bar{x_t},h_{t-1})$.	
			
			Sample $\left\lbrace z_t^{(i)}\right\rbrace \sim P(z_t)$.
			
			Sample $\left\lbrace \tilde{z}_t^{(i)}\right\rbrace\sim Q(z_t)$.
			
			Update hidden state $h_t \leftarrow f_\theta (h_{t-1},\left\lbrace \bar{x}_{t-1}^{(i)}\right\rbrace)$				
		}	
		
		$\left\lbrace z^{(i)}\right\rbrace = [\left\lbrace z_1^{(i)}\right\rbrace,...,\left\lbrace z_T^{(i)}\right\rbrace  ]$;
		$\left\lbrace \tilde{z}^{(i)}\right\rbrace=[\left\lbrace \tilde{z}_1^{(i)}\right\rbrace,...,\left\lbrace \tilde{z}_T^{(i)}\right\rbrace]$.
		
		Compute the adversarial discrimination loss: 
		$-\mathcal{L}_{dis}=\sum_{i=1}^{m} [D_\eta(\tilde{z}^{(i)})-D_\eta(z^{(i)})]$
		
		$g_\omega,g_\tau \leftarrow \nabla_{\omega,\tau}(-\mathcal{L}_{dis})$
		
		$\omega,\tau \leftarrow RMSProp(g_\omega,g_\tau )$
		
	\end{algorithm}
\end{figure}

\section{Theoretical analysis}
In the following analysis, we use the same notations in Algorithm \ref{alg1} and Algorithm \ref{alg2}. We denote the ELBO for VRNN in (\ref{elbo}) as $-\mathcal{L}_{ELBO}$. We define set $A$ and $B$ in the latent variable space $\mathcal{Z}x\mathbb{R}^d$: $A=\left\lbrace z|P(z)>0\right\rbrace$, $B=\left\lbrace z|Q(z)>0\right\rbrace$.

\begin{theorem}
	Given the assumption:
	\begin{itemize}
		\item $D$, $TR$ and $PS$ have enough capacity
		\item $D$ is allowed to achieve its optimum given $TR$ and $PS$.
	\end{itemize}
	
	When adversarial discrimination loss $-\mathcal{L}_{dis}$ achieves $0$, we have $\mathcal{L}_{ELBO}=\mathcal{L}_{rec}$. 
	
\end{theorem}

\begin{IEEEproof}
	
	Since $D$ has enough capacity, when $D$ achieves its optimum at each discriminator training step, $\mathcal{L}_{dis}$ is at minimum. With large training data samples $m$, according to the law of large numbers, we have:
	
	\begin{align}
		\begin{split}
			&\min_\eta \mathcal{L}_{dis}=\min_\eta \sum_{i=1}^{m} [D_\eta(z^{(i)})-D_\eta(\tilde{z}^{(i)})]\\
			&=m*\max_\eta \mathbb{E}_{z\sim Q(z_{\leq T})} [D_\eta(z)]-\mathbb{E}_{z\sim P(z_{\leq T})} [D_\eta(z)]\\
			&=m*W(P(z_{\leq T}),Q(z_{\leq T})).
		\end{split}
		\label{equ:29}
	\end{align}

	$W()$ is Wasserstein distance, i.e. EM (Earth-Mover) distance. This means before adversarial training step starts, $\mathcal{L}_{dis}$ becomes EM distance.
	
	Since $TR$ and $PS$ have enough capacity, when the adversarial discrimination loss $-\mathcal{L}_{dis}$ achieves $0$ at adversarial training step, we have:
	
	\begin{equation}
		-\mathcal{L}_{dis}=-m*W(P(z_{\leq T}),Q(z_{\leq T}))=0,
	\end{equation}
	\begin{equation}
		\Rightarrow W(P(z_{\leq T}),Q(z_{\leq T}))=0.
	\end{equation}
	
	According to \cite{arjovsky2017wasserstein}, when EM distance is $0$, we have the prior and posterior distribution at optimum distance:
	
	\begin{equation}
		P(z_{\leq T})=Q(z_{\leq T}).
		\label{eq17}
	\end{equation}
	
	We have the ELBO in unfactorization form in the right side of (\ref{elbo1}):
	
	\begin{align}
		\begin{split}
			&\mathcal{L}_{ELBO}=\mathbb{E}_{q(z_{\leq T}|x_{\leq T})} [log \frac{p(x_{\leq T},z_{\leq T})}{q(z_{\leq T}|x_{\leq T})}]\\
			&=\mathbb{E}_{q(z_{\leq T}|x_{\leq T})} [log \frac{p(x_{\leq T}|z_{\leq T},x_{<T})p(z_{\leq T}|z_{<T},x_{<T})}{q(z_{\leq T}|x_{\leq T})}].
		\end{split}
	\end{align}
	
	As in (\ref{eq17}), we have $p(z_{\leq T}|z_{<T},x_{<T})=q(z_{\leq T}|x_{\leq T})$. Then we have

	\begin{equation}
		\mathcal{L}_{ELBO}=\mathbb{E}_{q(z_{\leq T}|x_{\leq T})} [log p(x_{\leq T}|z_{\leq T},x_{<T})].
	\end{equation}
	
	As $\mathcal{L}_{rec}$ is the minus log likelihood of joint probability of each $x_t$, we have:
	
	\begin{equation}
		\mathcal{L}_{rec}=\sum_{t=1}^{T}\mathcal{L}_{rec}^t=\sum_{t=1}^{T}\mathbb{E}_{q(z_{\leq t})} [-log p(x_t)]=\mathcal{L}_{ELBO}.
	\end{equation}
\end{IEEEproof}

Theorem 1 essential says that when the adversarial regularization loss is minimized to 0 though updating on parameters of $TR$ and $PS$, the reconstruction phase training in Algorithm \ref{alg1} is equivalent to optimization on ELBO.

\section{Experiments}

\begin{figure*}[!h]
	\centering
	\fbox{\includegraphics[height=8.5cm]{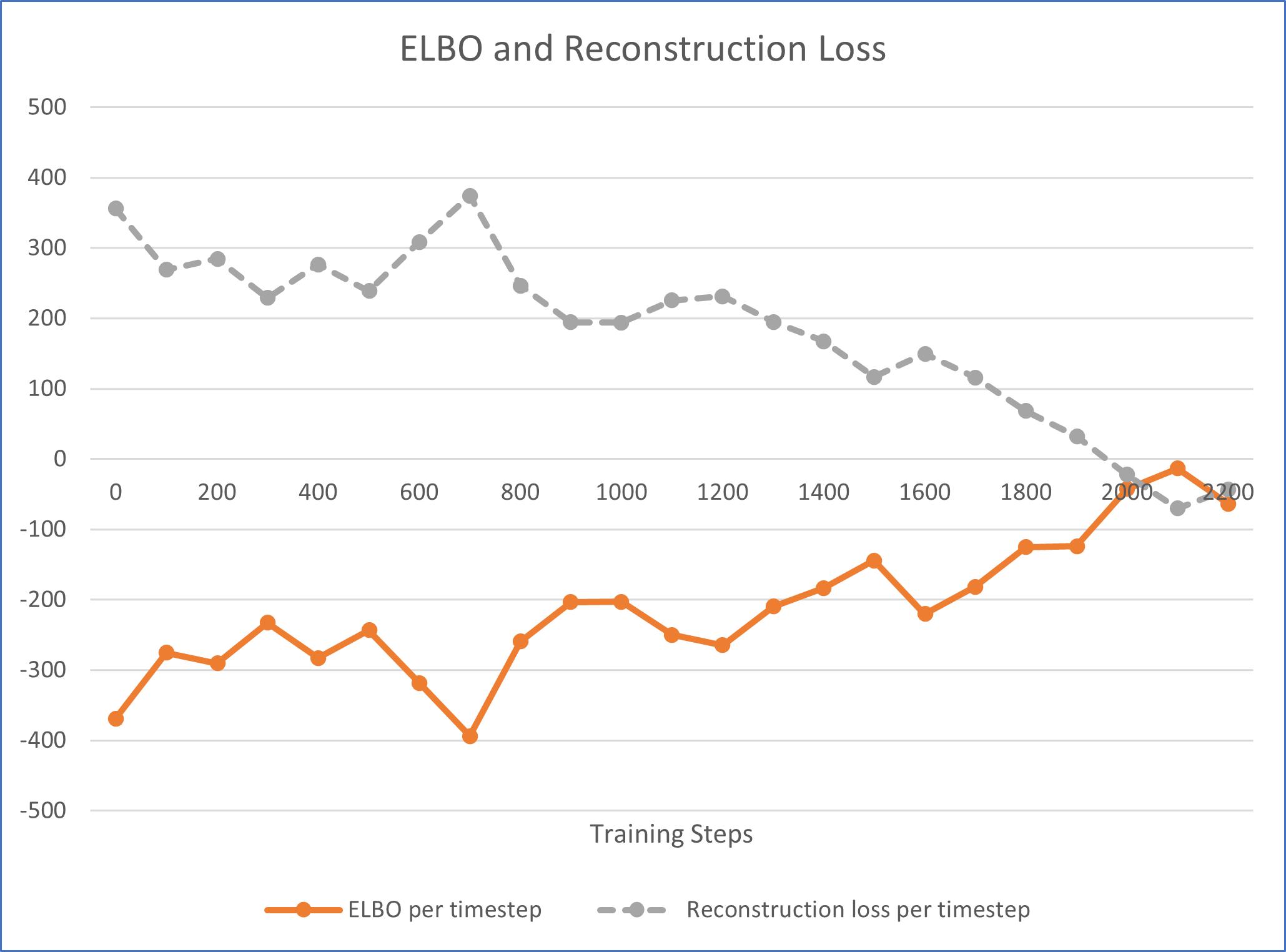}}
	
	\caption{ELBO and Reconstruction Loss converge to same level}
	\label{fig:loss1}
\end{figure*}

\begin{figure*}[!h]
	\centering
	\fbox{\includegraphics[height=6cm]{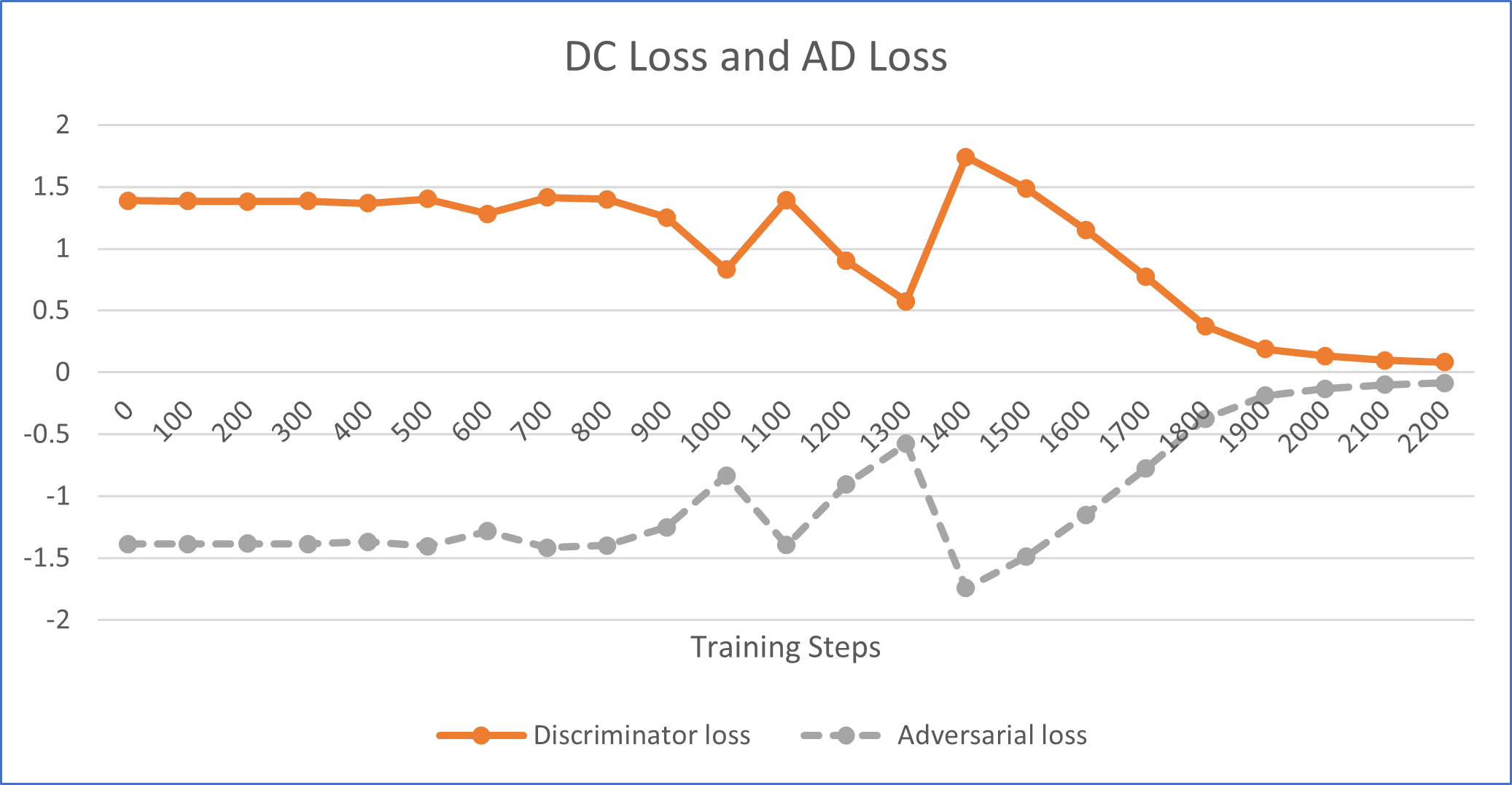}}
	
	\caption{Discriminator loss and adversarial loss approach to 0, with the training steps in Algorithm 2.}
	\label{fig:loss2}
\end{figure*}

We choose TIMIT speech data in our experiment \cite{garofolo1993darpa}. For both transition and emission model, we use the conditional normal distribution model. The sigma and mu of these conditional normal distributions are factorized by MLP neural networks. For proposal model, we use the normal approximated posterior distribution, similarly, factorized by MLP neural networks. The A-VRNN model includes a basic RNN model with trainable parameters in the RNN cell. The inputs to the RNN are the encoded latent variable $z$ of previous step as well as the encoded observed data point of current step. We use fully connected neural networks for the data encoder and latent encoder in this A-VRNN model. The data encoder accept the data points $x$ and encode it with desired dimensions before used as input to the RNN cell. Similarly, the latent encoder accepts $z$ and encodes it before used as input to RNN. 

For the discriminator $D$, one layer of LSTM and two layers of fully connect neural networks (FNET) are stacked together \cite{hochreiter1997long}. Hyper parameters are tuned with LSTM state size set as $128$ and number of nodes in the FNET set as $100$.

The experimental results are shown in Fig. \ref{fig:loss1} and \ref{fig:loss2}. These numerical results have validated the \textbf{Theorem 1}. Fig. \ref{fig:loss1} shows the decreasing reconstruction loss and increasing ELBO bound of the observed data points during the training process. The loss and bound results are summarized every $100$ training steps and averaged by length of the training sample sequences. The largest log likelihood per time-step (ELBO per time step) achieves $-13.3$ during the training.  Fig. \ref{fig:loss2} shows the converged discriminator loss and adversarial loss. Adversarial loss is the opposite of discriminator loss and is the EM distance of $P$ and $Q$.

\section{Discussion}

In AAE model, $z$ prior is know as GMM distribution, it is know prior. in A-VRNN, z prior is conditional distribution parameterized by the transition model, ie. TR neural networks in Fig. 1.   The conditional normal distribution prior presented by the TR gives A-VRNN the capability to model the dependency between adjacent latent variables. The objective of the adversarial training in both AAE and A-VRNN is to minimize the distance between prior distribution and posterior distributions of $z$.  In AAE model, this distance is measured with Jensen-Shannon Divergence, the same as the original GAN model \cite{goodfellow2014generative}. With this distance measurement, it is hard to achieve Nash equilibrium and unstable training process due to gradient vanishing \cite{salimans2016improved}. With our approach, the A-VRNN model use Wasserstein distance $W(P(z_{\leq T}),Q(z_{\leq T}))$ to achieve better training stability. Furthermore, our approach optimizes the trainable parameters in both factorized prior and posterior distribution model, i.e. TR and PS. This is to adapt to the different $z$ prior comparing to the fixed known prior in AAE. The experiment results show that this approach can achieve convergence within $2200$ training steps, while the approach used in AAE cannot converge using only the proposal loss.

\section{Conclusions}
We propose a novel training approach that addresses problems in regularization of the sequential latent variable model. Our approach use adversarial training in the model to regularize the latent variable distributions. Comparing to state of the art, our approach have below advantages:

\begin{itemize}
	\item Our approach has achieves optimum in the training algorithms and provides better model training robustness;
	\item Our approach improves the posterior approximation and has a clear track of the factorization of the posterior distribution;
	\item The symmetric EM distance used in our approach provides smooth measure of latent variable distribution distance between prior and posterior. This gives better training stability;
\end{itemize}

\bibliographystyle{IEEEtran}
\bibliography{IEEEabrv,bibfile.bib}

\end{document}